\documentclass[11pt]{article}
\usepackage{coling2018}
\usepackage{times}
\usepackage{color}
\usepackage{latexsym}
\usepackage{multirow}
\usepackage{graphicx} 
\usepackage{amssymb}
\usepackage{amsmath}
\usepackage{url,comment}
\usepackage{natbib}
\usepackage{float}
\usepackage{fancyhdr}

\makeatletter
\newcommand{\@BIBLABEL}{\@emptybiblabel}
\newcommand{\@emptybiblabel}[1]{}
\makeatother
\usepackage{hyperref}

\title{Joint Neural Entity Disambiguation with Output Space Search}

\author{Hamed Shahbazi, Xiaoli Z. Fern, Reza Ghaeini, Chao Ma, \\\textbf{Rasha Obeidat, Prasad Tadepalli}
\\
Oregon State University, Corvallis, OR, USA \\
\{shahbazh, xfern, ghaeinim, machao, obeidatr, tadepall\}@eecs.oregonstate.edu
}
\date{}

\begin{document}
\maketitle
\begin{abstract}
In this paper, we present a novel model for entity disambiguation that combines both local contextual information and global evidences through Limited Discrepancy Search (LDS). Given an input document, we start from a complete solution constructed by a local model and conduct a search in the space of possible corrections to improve the local solution from a global view point. Our search utilizes a heuristic function to focus more on the least confident local decisions and a pruning function to score the global solutions based on their local fitness and the global coherences among the predicted entities. Experimental results on CoNLL 2003 and TAC 2010 benchmarks verify the effectiveness of our model.
\end{abstract}
\section{Introduction}
\blfootnote{
%
% for review submission
%
% \hspace{-0.65cm}  % space normally used by the marker
% Place licence statement here for the camera-ready version. See
% Section~\ref{licence} of the instructions for preparing a
% manuscript.
%
% % final paper: en-uk version 
%
% \hspace{-0.65cm}  % space normally used by the marker
% This work is licensed under a Creative Commons 
% Attribution 4.0 International Licence.
% Licence details:
% \url{http://creativecommons.org/licenses/by/4.0/}.
% 
% % final paper: en-us version 
%
\hspace{-0.65cm}  % space normally used by the marker
This work is licensed under a Creative Commons Attribution 4.0 International License. License details: \url{http://creativecommons.org/licenses/by/4.0/}.}
The goal of entity disambiguation is to link a set of given query mentions in a document to their referent entities in a Knowledge Base (KB). As an essential and challenging task in Knowledge-Base Population (KBP) for text Analysis
\citep{tac:ji14,tac:ji15}, entity disambiguation has attracted many research efforts from the NLP community. Recently, deep learning based approaches have demonstrated strong performances on this task \citep{rel:hoffman,rel:ibm}.

A main challenge for entity disambiguation is to best identify and represent the appropriate context, which can be local or global. Different methods have been proposed to capture and represent different types of contexts. Textual context has been heavily investigated for local models that score each query's candidates independently. Representations of the textual contexts range from weighted combination of the word embeddings based on attention \citep{rel:hoffman}, to more fine-grained contextual representations using recurrent neural networks \citep{rel:ibm}. The global context of other entities in the document has also been studied for a more global and joint prediction view on the problem. \cite{rel:hoffman} use a Conditional Random Field (CRF) based model to capture the interrelationship among entities in the same document, whereas \cite{rel:google} introduce a soft k-max attention model to weigh the importance of other entities in the document in making prediction for any given query.

Working with the recently proposed models, we observe that local models that employ an appropriate attention mechanism often have a solid linking performance. In a single document, there are often a small number of hard queries for which the local model fails to make a correct decision. We conjecture that if some of these mistakes can be corrected, a global model that enforces coherence among entities will be able to propagate these corrections to improve the overall solution quite effectively.

This inspires us to consider the Limited Discrepancy Search (LDS) framework \citep{JMLR:v15:doppa14a}, which conducts a search over possible corrections on a complete output with the goal of improving the final output. Critically, LDS works well for cases where only a small number of local corrections are needed to reach a good global solution. This nicely matches up with our observation of the behavior of entity disambiguating models. 

In this paper, we propose a LDS based global entity disambiguating system. Given a document and its query mentions, our system first applies a local disambiguation model to produce an initial solution. We then use LDS to conduct a shallow search in the space of possible corrections (with the focus on hard/least confident mentions) to find a better solution. Evaluation on CoNLL 2003 and TAC 2010 shows that our method outperforms the current state-of-the-art models. We also conduct an extensive ablation study on different variants of our model, providing insight into the strength of our method.

\section{Proposed Approach}
%\subsection{Problem Formulation} 
We are given a document $D$ containing $n$ mentions $[x_1, ... x_n]$. We assume that all mentions are linkable to a Knowledge Base (KB) by excluding the NILL mentions. We are interested in finding a joint assignment of all mentions to $\textbf{Y}=[y_1, ... y_n]$ of referent entities to maximize the following score:
\begin{equation}
s(\textbf{Y}) = \sum_{i=1}^{n}\psi(x_i, y_i) + \sum_{i=1}^{n}\sum_{j=1:j\neq i}^{n}\phi(y_i, y_j)
\label{map}
\end{equation}
where function $\psi(x_i, y_i)$ gives the local compatibility score between the mention $x_i$ and its candidate $y_i$ and $\phi(y_i, y_j)$ indicates the amount of relatedness between the assigned candidates $y_i$ and $y_j$. Optimizing this objective, however, is NP-hard. In this work, we develop a LDS-based search strategy for optimizing this objective.\\

\subsection{Overview of the Approach}
Given a document with $n$ mentions, we initialize the search with a solution acquired based solely on the local scoring function $\psi(\cdot, \cdot)$. We then conduct
a greedy beam search in the space of possible
discrepancies (changes/corrections) to this initial solution while focusing on mentions with least confident local scores in the hope of finding a better solution.
\begin{figure*}[h]
\centering
\includegraphics[width=0.94\textwidth,height=0.40\textwidth]{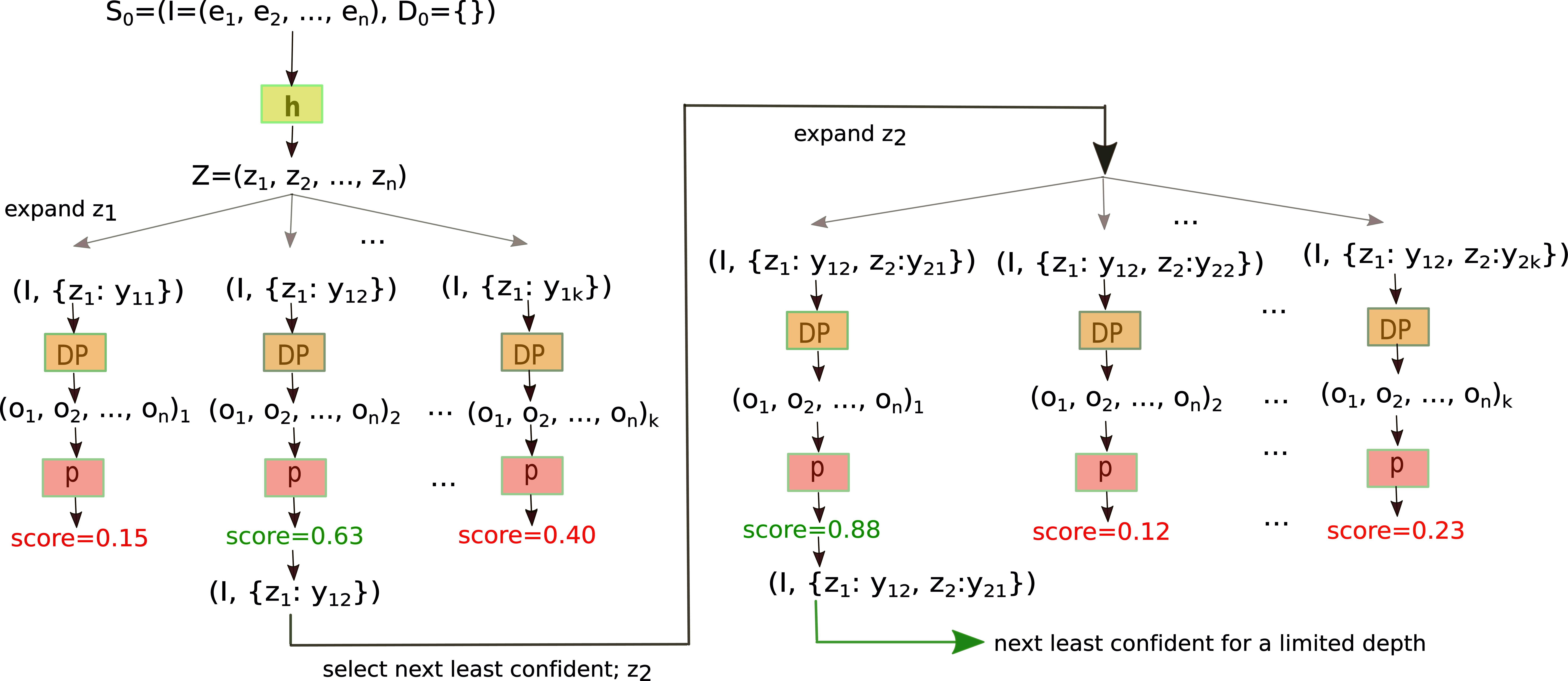}
\caption{\label{fig:general_model} The general framework of our proposed model with beam size $b$=1.}
\vspace{-0.15in}
\end{figure*}
Fig~\ref{fig:general_model} shows the overview of our search framework for beam size $b = 1$. Each state $S_i$ is a pair $(I, D_i)$ where $I = (e_1,e_2,...,e_n)$ is the initial solution given by the local model and $D_i$ is the discrepancy set for state $S_i$. For example, a discrepancy set $\{x_{k_1} :y_{k_1}, x_{k_2} :y_{k_2} \}$  contains two discrepancies, changing the assignment for mentions $x_{k_1}$ and $x_{k_2}$ from $e_{k_1}$ and $e_{k_2}$ in $I$ to $y_{k_1}$ and $y_{k_2}$ respectively. 
Starting with initial state (I, {}), we utilize a heuristic function $h$ (Section~\ref{sec:h}) to sort the mentions in the increasing order of their local confidence. Let $(z_1,...,z_n)$ be the ordered mentions where $z_1$ is the least confident mention. Each iteration of the greedy beam search explores and prunes the space of discrepancy sets as follows:

We select the least confident mention $z_1$ and expand state $(I, \{\})$ to $k$ new
states $(I, \{z_1 : y_{1j}\})$ for $j = 1 \cdots k$ where $y_{1j}$
is the $j^{th}$ candidate for mention $z_1$ (we consider top $k$ most probable candidates). Each expanded
state $(I, \{z_1 : y_{1j}\}), j=1,\dots,k$ is given to a Discrepancy Propagator ($DP$) (Section~\ref{sec:dp}) to get
its discrepancy set propagated throughout the document and produce an updated complete solution $(o_1, ... o_n)_j$, for $j=1,\dots,k$.
%The $DP$ is trained
%and utilized to reduce the depth of the search
%by propagating the discrepancies to the document.
%Given a state to the $DP$ it
%$DP$ propagates the discrepancy set and returns a updated complete solution $(o_1, ... o_n)_k$. Therefore for $k$ expanded states we produce $k$ complete solutions. 
Utilizing a trained pruning function $p$  (Section~\ref{sec:p}) we then rank the $k$ complete solutions and prune them
to top $b$ states ($b$ is beam size).

The search continues with the next least confident mention $z_2$ as
shown in Fig~\ref{fig:general_model}. Each iteration increases the size
of the discrepancy set by one. Note that a mention
is not repeated in a discrepancy set.\\
We consider two different
strategies for terminating the search depending
on the used heuristic function $h$ (Section~\ref{sec:p}).  Each strategy causes the search to terminate at different depth (length) of the discrepancy set. The output of the search is selected by the pruning function $p$ from the last set of complete solutions.\\
In the following, we will first explain our \textbf{local model} ($\psi(\cdot,\cdot)$) for producing the initial solution. We will then introduce our LDS search framework, and its key components including the \textbf{Discrepancy Propagator} to propagate a set of discrepancies to other mentions in the document; the \textbf{Heuristic Function} to compute the local confidence of the mentions in the document and identify the least confident ones; and finally the \textbf{Pruning Function} to guide the search and select the final solution.

%The key elements of our LDS framework include
%the \textbf{Local Model} ($\psi(\cdot,\cdot)$) to produce the initial solution; the \textbf{Heuristic Function} to compute the local confidence of the mentions in the document and identify
%the least confident ones; the \textbf{Discrepancy Propagator} to propagate a set of discrepancies to other mentions in the document and finally the \textbf{Pruning Function} to guide the search. Below we describe each element in detail.
\subsection{Local Model}
\label{sect:local}
Our local model utilizes contextual, lexical and prior evidences to compute the compatibility score $\psi(x_i, y_i)$ for assigning candidate $y_i$ to mention $x_i$. These evidences are extracted and used as follows: 
\subsubsection{Contextual Evidence}
\label{subsec:contextual}
Given a query mention, a key challenge for the local model is to identify a minimum but sufficient contextual evidence to disambiguate the query. We first extract all sentences in the document relevant to the query mention. This is achieved by applying CoreNLP \citep{l:corenlp} to perform coreference resolution to all mentions in the document and extract all sentences containing a mention in the query's coreference chain. 

The set of sentences are then concatenated to form the context for the query; $w_i=[w_{i1}, ... w_{im}]$, where $w_{ij}\in R^d$ is the embedding of the $j$-th word in the context. We then use an attention model introduced by \citep{rel:hoffman} to compress the context into a single embedding. Specifically, we define the contextual representation $c_i\in R^{d}$ for mention $x_i$ with candidate set $\{y_{i1}, y_{i2}, ... y_{ik}\}$ over context $w_i$ as 
\begin{equation}
c_i = \sum_{l=1}^{m} \alpha_{il} w_{il}
\end{equation}
The weight vector $\alpha$ is computed using the following attention considering all $k$ entity candidates: 
\begin{equation}
\alpha_i = \textbf{softmax} ([\max_{j \in 1..k} y_{ij}^{T}Aw_{i1}, ..., \max_{j \in 1..k}y_{ij}^{T}Aw_{im}])
\end{equation}
where $A$ is a learned matrix that scores the relatedness between word and entity.
This attention model computes the relatedness of each word with all of the entity candidates and takes the max as the score for each word. The scores of all context words are then passed through softmax to compute their weight. Under this model, if a word is strongly related to one of the candidates, it will be given a high weight. Subsequently, using $c_i$ we define the following contextual features for candidate $y_{ij}$ as $f^{(c)}_{ij}=[y_{ij}^T B c_{i}; y_{ij}^T B; c_i]$, where $B$ is a learned $R^d \times R^d$ matrix. Note that $f^{(c)}_{ij} \in R^{2d+1}$. 

\subsubsection{Lexical Evidence}
\label{subsec:lexical}
The contextual features extracted above ignores any lexical/surface information between query mention and entity title, which can be useful. To this end we include some lexical features to our local model including variants of the edit distance between the surface strings of the mention and the candidate title, whether their surface strings follow an acronym pattern and etc. Detailed list can be found in the Table~\ref{tab:LFs}(a).
The extracted features are scalar real values. We use RBF binning \citep{rel:ibm} to transform each scalar to a $10$-d vector. Hence for each query $x_i$ and candidate $y_{ij}$ we have lexical vector feature $f_{ij}^{(l)}\in R^{10 \times |f|}$ where $|f|$ is the number of features listed in Table~\ref{tab:LFs}(a).
\begin{table}[h!]
\begin{minipage}[b]{0.50\linewidth}
\centering
(a)\\
\small
\begin{tabular}{|l|}
\hline \textbf{mention:} m = [$m_1$, ... $m_a$],  \textbf{entity title:} e = [$t_1$, ... $t_b$] \\
\hline
$f_1$:\hspace{1mm} mention length = $a$\\
$f_2$:\hspace{1mm} entity title length = $b$\\
$f_3$:\hspace{1mm} $\sum_{i=1}^{b}$ (occurrence counts of $t_i$ in the document)\\
$f_{4}$:\hspace{1mm} m is acronym\\
$f_{5}$:\hspace{1mm} e is acronym\\
$f_{6}$:\hspace{1mm} m and e acronym patterns are exact match\\
$f_{7}$:\hspace{1mm} m and e are non-acronyms and exact match\\
$f_{8}$:\hspace{1mm} min-edit(m, e)\\
$f_{10}$:\hspace{1mm} sum of partial min edits:\\ \hspace{10mm} $\sum_{i=1}^{a}(\min_{j=1}^{b}(\mbox{min-edit}(m_i, t_j)))$\\
\\
\hline
\end{tabular}
\end{minipage}
\hspace{0.01cm}
\begin{minipage}[b]{0.40\linewidth}
\centering
(b)\\
\begin{tabular}{|l|}
\hline local score for mention $m$:\\ $l$ = \textbf{sotmax}([$\psi(x_i, y_{i1})$, ... $\psi(x_i, y_{ik})$])\\
\hline
$f_1$:\hspace{1mm} max($l$)\\
$f_2$:\hspace{1mm} second-max($l$)\\
$f_3$:\hspace{1mm} entropy($l$)\\
$f_{4}$:\hspace{1mm} $m$ is an acronym\\
$f_{5}$:\hspace{1mm} length($m$)\\
\\
\\
\hline
\end{tabular}
\end{minipage}
\hspace{0.01cm}
\begin{minipage}[b]{0.30\linewidth}
\centering
\end{minipage}
\caption{(a) list of lexical features in the local model (b) list of the features to learn heuristic function $h_2$}
\label{tab:LFs}
\end{table}
\vspace{-0.15in}
\subsubsection{Prior Evidence}
\label{subsec:prior}
We consider $p(e|m)$, the prior probability that an entity $e$ is linked to a mention string $m$ as prior evidence. This is computed using hyper-link statistics from Wikipedia and aliases mapping from \citep{l:hoffart} obtained by extending the “means” tables of YAGO \citep{l:hoffart13}. The $p(e|m)$ is also transformed to a $10$-d vector using RBF binning \citep{rel:ibm} to create prior feature $f_{ij}^{(p)} \in R^{10}$.

\subsubsection{Overall local model}
The contextual, lexical and prior features are concatenated and fed through a Multi Layer Perceptron (MLP) with 2 hidden layers (relu, 200-d and 50-d respectively) to produce a final local score for all candidates as follows:
\begin{equation}
\psi(x_i, y_{ij}) = \sigma(W_s[f_{ij}^{(c)}; f_{ij}^{(l)}; f_{ij}^{(p)}] + b_s)
\end{equation}
Where $W_s$ and $ b_s$ are weight and bias parameters for the MLP. The learning of the model is two tiered. We first pre-train the matrices $A$ and $B$ using the cross-entropy loss by using $\textbf{softmax}([y_{ij}^{T}Bc_{i}]_{j \in 1..k})$ as the predicted probability for each candidate for mention $x_i$, 
Keeping $A$ and $B$ fixed, we then train the MLP weights with a drop-out rate of 0.7 minimizing again the cross-entropy loss. Note that we use the entity embeddings produced by \citep{rel:hoffman} and the word embeddings produced by \citep{l:Mikolov} using skip-gram model.

\subsection{Global Model via LDS}
Our local model primarily focuses on the textual context of each mention in making its decisions and ignores the relationship between mentions. Our global model takes as input the local predictions for all the mentions in a single document and constructs a globally coherent final solution using Limited Discrepancy Search. Before we introduce our LDS search procedure, we will first describe the discrepancy propagator which is critical toward achieving an efficient search space.

\subsubsection{Discrepancy Propagator}
\label{sec:dp}
The purpose of the discrepancy propagator is to allow the influence of the local changes to propagate to other parts of the solution, thus reducing the necessary number of discrepancies needed.  Our discrepancy propagator is essentially a new scoring function that evaluates each candidate for a given mention not only based on the local compatibility but also the global coherence among the predictions. 

Given that not all mentions in a document need to be related to one another, for each mention $x_i$ we construct an entity context $E_i$ considering a window of 30 mentions centered at query mention $x_i$. Given the current entity assignment $e_1, e_2, ... e_n$ to the mentions in the document, the entity context for mention $x_i$ will be $E_i=e_{i-15} ... , e_{i-1}, e_{i+1}, ... e_{i+15}$.\\
For a given mention $x_i$ and its entity context $E_i$, the new score of a candidate $y_{ij}$ is defined as:
\begin{equation}
\label{global}
g(x_i, y_{ij}) = \psi(x_i, y_{ij}) + \frac{1}{|E_i|}\sum_{e\in E_i} \phi(e, y_{ij})
\end{equation}
In equation~\ref{global}, the score of a candidate now considers both its local score $\psi$ as well as its average coherence with the other predictions in its entity context. The coherence score $\phi$ between two entities $e$ and $y_{ij}$ is defined as:
\begin{equation}
\label{compatibility}
\phi(e, y_{ij}) = e^TCy_{ij} + w^T r(e, y_{ij})
\end{equation}
where $e$ and $y_{ij}$ are entity embeddings and $r(e, y_{ij})$ is a vector of three pairwise features between the two entities: log transformed counts of co-ocurrences, shared in-links and shared out-links; $C \in R^{d\times d}$ and $w\in R^3$ are learned weights. \\
The score given by equation~\ref{global} is fed through a softmax activation to assign probabilities to each candidate of $x_i$. We train $W$ and $c$ in a mention-wise procedure using cross entropy loss. Specifically for each mention $x_i$ we use ground truth entities for the other mentions in $E_i$ and update weights of $C$ and $w$ such that the true candidate of $x_i$ gets higher score than its false candidates.\\
Note that we do not intend to use this scoring function to do joint inference on all mentions. Instead, it is used with LDS to propagate the discrepancies to the output of all related mentions as follows. Given the original solution $e_1, e_2, ... e_n$ and a set of discrepancies $\{x_k:y_k\}$, we first update $e_k$ to $y_k$. Then we re-evaluate equation~\ref{global} for all mentions that have $y_k$ in their entity context to produce a complete solution based on new entity context containing $y_k$. 
\subsubsection{Heuristic Function}
\label{sec:h}
The heuristic function takes the initial solution and sort all the mentions based on their prediction confidence. In this work we consider two heuristic functions for computing the prediction confidence of the local model. For a given query $x_i$, our first heuristic function $h_1$ computes its confidence simply by taking the max of the local scores normalized by a softmax funtion:
\begin{equation}
h_1(x_i) = \textbf{max}(\textbf{softmax}([\psi(x_i, y_{i1}), ...\psi(x_i, y_{ik})]))
\end{equation} 
where $\psi(x_i, y_{ij})$ is the local score for candidate $y_{ij}$. 

For our second heuristic function $h_2$, we learn a binary classifier (a two layer MLP) to produce local confidence for each mention $x_i$. The binary classifier utilizes the features listed in Table~\ref{tab:LFs}(b). Each feature value is transformed into a $10$-d vector using RBF binning \citep{rel:ibm}. The training samples for the binary classifier are generated from the prediction of the local model on the training set. One training example is generated for each local prediction --- if the local model is correct about its decision then the label for the sample is 1. In order to balance the positive and negative training samples, we randomly sub-sample positive local predictions. For each local prediction, the learned classifier outs a probability of its being correct, which is used as the confidence score.
\subsubsection{Pruning Function}
\label{sec:p}
At each iteration of the search, each beam is expanded to $k$ new states. Following the expansion, the DP is applied to the $b\times k$ expanded nodes and re-evaluates equation~\ref{global} for all mentions with updated entity contexts due to the discrepancies. It hence propagates the discrepancies and produces a complete solution. Subsequently, the pruning function evaluates each of the $b\times k$ complete solutions and reduces the expansion list to the beam size $b$.

The pruning function is similar to equation~\ref{global} but scores all the mentions collectively. Given all the mentions in the document $x_1, ..., x_n$ and their predicted entities denoted as $o = o_1, ..., o_n$, our pruning function scores the solution $o$ as follows:
\begin{equation}
s(o) = \sum_i \psi(x_i, o_i) + \sum_{(o_i\in E_j \mbox{ or } o_j \in E_i)} \phi_g(o_i, o_j)
\end{equation}

Where constraint $(o_i\in E_j \mbox{ or } o_j \in E_i)$ indicates that we consider relation among pair of entities $(o_i, o_j)$ if they are in the entity context of one another. Here $\phi_g$ takes the same form as equation~\ref{compatibility} but uses a different set of weights $C_g$ and $w_g$. 

Despite the similarity in forms, the pruning function serves a different purpose from that of equation~\ref{compatibility} and requires different training. We train the pruning function by reducing it to a rank learning problem. Specifically, in training, we collect all the complete solutions that are considered in each pruning step, and compute their hamming losses from the ground truth. Given a set of solutions in a pruning step, we will create ranking pairs  to require the solution with the least hamming loss to score higher than all the others. Given a ranking pair, $o^t$ and $o^f$, assuming $o^t$ has lower hamming loss than $o^f$, we use the following ranking loss for training: 

\begin{equation}
max\left(0, \Delta(o^{t}, o^{f}) - s(o^{t}) + s(o^{f})\right)
\label{eq:hinge_rank}
\end{equation}
where $\Delta(o^{(t)}, o^{(f)})$ is the absolute difference of the hamming-loss between $o^{(t)}$ and $o^{(f)}$. This loss function penalizes the scoring function if it fails to score $o^t$ higher than $o^f$ by a margin specified by $\Delta(o^{(t)}, o^{(f)})$.

\paragraph{Terminating the search.}
We consider two different strategies for terminating the search depending on the heuristic function in use. When using $h_1$ as the heuristic function, search terminates when we reach a depth limit $\tau$ (a maximum $\tau$ discrepancies). The strategy with $h_2$ uses a flexible depth. For this strategy, we terminate the search when discrepancies have been introduced for all queries that are predicted to be incorrect by $h_2$. 

\section{Experiments}
\label{sec:exp}

\subsection{Data Sets}
We use two datasets CoNLL 2003 \citep{l:hoffart} and TAC 2010 \citep{l:ji10} for evaluation. The CoNLL dataset is partitioned into train, test-a and test-b with 946, 216 and 231 documents respectively. Following our baselines we only use 27816 mentions with valid links to the KB. The TAC 2010 dataset is yet another popular NED dataset released by the Text Analysis Conference (TAC). The dataset contains training and test set with 1043 and 1013 documents respectively. Similar to CoNLL we only consider linkable mentions in TAC and report our performance on 1020 query mentions in the test set.\\
To learn and tune the parameters of the local models for CoNLL and TAC we use their own training and development splits. However, to learn and tune the parameters of the global model (the discrepancy propagator, the heuristic and pruning functions) we only use the CoNLL training and dev-sets.\\
The number of queries per document in the test sets of the CoNLL and TAC are approximately 20 and 1 respectively. In order to have a global setup for TAC we apply CoreNLP \citep{l:corenlp} mention extractor to the test documents of TAC and perform joint disambiguation of the extracted mentions together with the query mentions, increasing the number of mentions to approximately 4 per document. We only report the performance on the query mentions with the given standard ground truth by TAC.
\subsection{Hyper-parameters and Dimensions} Our model settings for the hyper-parameters and dimensionality of the embeddings and weights are as follows: We use entity/word embeddings of 300-d. We learned the embeddings using the mechanism proposed by \citep{rel:hoffman}. We use 2-layers MLPs for the local model and the heuristic $h_2$ with hidden layer sizes 200$\times$50 and 100$\times$20 for each respectively. The RBF binning always transfers a scalar to 10-d.  Although we analyze and compare different configurations for beam-size and depth limit in section~\ref{section:ablation}, we use beam size 5 and flexible depth limit $\tau$ with heuristic $h_2$ in our reported results. 
\subsection{Results}
To evaluate the model performance we use the standard micro-average accuracies of the top-ranked candidate entities. We use different alias mappings for TAC and CoNLL. Specifically, for TAC we only use anchor-title alias mappings constructed from hyper-links in the Wikipedia. For CoNLL, in order to follow the experimental setup reported by \citep{rel:ibm}, in addition to the alias mappings of Wikipedia anchor-titles, we use the mappings produced by \citep{l:pershina} and \citep{l:hoffart}.\\
Tables~\ref{tab:TAC}(a) and (b) show our performance on the CoNLL and TAC datasets for our local and global models along with other competitive systems respectively. The prior state of the art performances on these two datasets are achieved by \citep{rel:ibm}. The results show that our global model outperforms all the competitors for both CoNLL 2003 and TAC 2010. It is interesting to note that our local model is solid, but is noticeably inferior to the stat-of-the-art local model. With an even stronger local model like that of \citep{rel:ibm}, one can potentially expect our LDS-based global model to push the state-of-the-art even further.
\begin{table}[h!]
\begin{minipage}[b]{0.45\linewidth}
\centering
\footnotesize
\begin{tabular}{|l|c|}
\hline \bf models & \bf In-KB acc\%\\
\hline
\multicolumn{2}{|l|}{\bf local} \\
\hline
\citep{l:He} &  85.6 \\
\citep{l:Francis} &  85.5 \\
\citep{l:sil:16} &  86.2 \\
\citep{l:lazic} &  86.4 \\
\citep{l:yamada} &  90.9 \\
\citep{rel:ibm} & 94.0 \\
\hline
\multicolumn{2}{|l|}{\bf global} \\
\hline
\citep{l:hoffart} &  82.5 \\
\citep{l:pershina} &  91.8 \\
\citep{rel:google} &  92.7 \\
\citep{l:yamada} &  93.1 \\
\hline
\multicolumn{2}{|l|}{\bf our model} \\
\hline
local &  90.89 \\
global & \textbf{94.44} \\
\hline
\end{tabular}\\ 
(a) CoNLL 2003
\end{minipage}
\hspace{0.01cm}
\begin{minipage}[b]{0.45\linewidth}
\centering
\footnotesize
\begin{tabular}{|l|c|}
\hline \bf models & \bf In-KB acc\%\\
\hline
\multicolumn{2}{|l|}{\bf local} \\
\hline
\citep{l:sil:16} & 78.6 \\
\citep{l:He} & 81.0 \\
\citep{l:sun} & 83.9 \\
\citep{l:yamada} & 84.6 \\
\citep{rel:ibm} & 87.4 \\
\hline
\multicolumn{2}{|l|}{\bf global} \\
\hline
\citep{l:yamada} & 85.2 \\
\citep{rel:google} & 87.2 \\
\hline
\multicolumn{2}{|l|}{\bf our model} \\
\hline
local & 85.73 \\
global & \textbf{87.9} \\
\hline
\end{tabular}\\
(b) TAC-2010
\end{minipage}
\hspace{0.01cm}
\vspace{-0.15in}
\caption{Evaluation on CoNLL 2003 Test-b and TAC-2010}
\label{tab:TAC}
\vspace{-0.25in}
\end{table}
\subsection{Ablation Study and Performance Analysis}
\label{section:ablation}
For ablation we analyze the impact of the features and search. We also analyze the behavior of our model based on the rarity of the entities.
\subsubsection{Feature Analysis}
We study the impact of features on the local and global models. For the local model, starting with only considering the contextual evidence as described in Section~\ref{subsec:contextual}, we see that the performance steadily increases as we add the prior and lexical features. As shown in tables~\ref{tab:TAC_F}(a) and (b) the prior and lexical features have very strong impact on TAC. The binning technique that projects the prior and lexical features to 10 dimensions gives an average of 0.94\% and 0.61\% percentage points for TAC and CoNLL respectively.\\
For the global model, entity pair compatibility is computed using both entity embeddings and log transformed counts of co-occurences, shared in-links and shared out-links. Using these features leads to a gain of 1.1\% and 0.43\% in accuracy for CoNLL and TAC respectively compared to the global model using only the embeddings in equation~\ref{compatibility}.
\begin{table}[h!]
\begin{minipage}[b]{0.45\linewidth}
\centering
\small
\begin{tabular}{|l|c|}
\hline \bf models & \bf In-KB acc\%\\
\hline
\multicolumn{2}{|l|}{\bf local} \\
\hline
Context only &  85.37 \\
Context + Prior + Lexical &  90.28 \\
Context + Prior + Lexical + Bin &  90.89 \\
\hline
\multicolumn{2}{|l|}{\bf global} \\
\hline
Our global & 94.44 \\
- log count features  &  93.34 \\
- LDS + 1-step global prop. &  93.14 \\
- LDS + conv. global prop. & 93.63 \\
% -LDS +LC -ITR &  93.14 \\
% +LDS -LC -ITR &  93.34 \\
% +LDS +LC -ITR & 94.44 \\
% -LDS +LC +ITR & 93.63 \\
\hline
%\multicolumn{2}{|l|}{\tiny LC: log counts} \\
%\multicolumn{2}{|l|}{\tiny LDS: search} \\
%\multicolumn{2}{|l|}{\tiny  ITR: iterative global  propagation until convergence} \\
%\hline
\end{tabular}\\
(a) CoNLL 2003
\end{minipage}
\hspace{0.01cm}
\begin{minipage}[b]{0.45\linewidth}
\centering
\small
\begin{tabular}{|l|c|}
\hline \bf models & \bf In-KB acc\%\\
\hline
\multicolumn{2}{|l|}{\bf local} \\
\hline
Context & 70.86 \\
Context + Prior + Lexical & 84.79 \\
Context + Prior + Lexical + Bin & 85.73 \\
\hline
\multicolumn{2}{|l|}{\bf global} \\
\hline
Our global & 87.9 \\
- log count features & 87.47 \\
- LDS + 1 step global prop. & 86.21 \\
- LDS + conv. global prop. & 86.29 \\

% -LDS +LC -ITR & 86.21 \\
% +LDS -LC -ITR & 87.47 \\
% +LDS +LC -ITR & 87.9 \\
% -LDS +LC +ITR & 86.29 \\

% \hline
% \multicolumn{2}{|l|}{\tiny LC: log counts} \\
% \multicolumn{2}{|l|}{\tiny LDS: search} \\
% \multicolumn{2}{|l|}{\tiny ITR: iterative global propagation until convergence} \\
 \hline
\end{tabular}\\
(b) TAC-2010
\end{minipage}
\hspace{0.01cm}
\vspace{-0.15in}
\caption{The performance of variants of our Local/Global models on CoNLL 2003 and TAC-2010}
\label{tab:TAC_F}
\vspace{-0.15in}
\end{table}
%In the local and global models we note no significant benefits when using square $A$, $B$ and $C$ to the diagonal ones.
\subsubsection{Search Analysis}
In the last two rows of Table~\ref{tab:TAC_F}, we list variants of our model where LDS search is removed. Specifically, given the initial local solution, our discrepancy propagator (equation~\ref{global}) can be applied without any discrepancy to re-evaluate the candidates for all mentions and obtain a more globally compatible solution. Naturally, one can repeat this for multiple iterations till convergence, which gives us an alternative global model that does not rely on search. We consider this model and show its single iteration performance (-LDS + 1-step global prop.) as well as its converged performance (-LDS + conv.~global prop.) in Table~\ref{tab:TAC_F}. From the results we can see that a single iteration of the global propagation was able to significantly improve the initial local solution, but later iterations lead to very mild gain. This is because after the first propagation the convergence is almost immediate and there is a limited improvement after the first iteration. In contrast, our search based global model is able to achieve significant further performance gain 1.3\% for CoNLL and 1.68\% for TAC. The reason is that with local discrepancies introduced by LDS, we force the solution to escape the current local optimum and search for better ones.

Additionally, several parameters/choices in the search can potentially impact the performance: the beam size $b$, the depth of search, and the different heuristics for prioritizing the discrepancy locations. The following set of experiments explore these choices of parameters:
\paragraph{Beam size.} We compare model with beam size $b=5$ and a simple greedy model with $b=1$. For CoNLL 2003 and TAC-2010, the model with $b=5$ gives a gain of 0.24\% and 0.13\% compared to the greedy. Increasing the size of the beam further beyond 5 did not show significant gain.
\paragraph{Depth of search.} For heuristic $h_1$, we use a fixed depth strategy. In particular, given a document with $n$ mentions, we consider two different depth limits: $\tau=25\% n$ and $\tau=50\%n$, which lead to an average depth of 5 and 10 per document respectively. For heuristic $h_2$, the depth is flexible and determined by the number of mentions that are predicted to be incorrect by $h_2$. This strategy leads to an average depth of $4$. In Tables ~\ref{tab:confusion} (a) and (b), we report the confusion matrices given by $h_1$ applied to the local prediction on test-b of CoNLL 2003 with $\tau=25\%n$ and $\tau=50\%n$ respectively. In these tables correct/incorrect mean whether the local prediction is correct or not (comparing to the ground truth). Therefore the first cell with value 1134 indicates that there are 1134 mentions in test-b that are correctly predicted by the local model, but deemed as among the top 25\% least confident mentions (aka the hard queries) by $h_1$. The confusion matrix for a good heuristic will have small diagonal values and large anti-diagonal values. 

In Table~\ref{tab:confusion} (c), we apply heuristic $h_2$ to the same data with flexible depth. These results show that heuristic $h_2$ gives the best precision as well as recall of the real mistakes made by the local model. 
\begin{table}[h!]
\begin{minipage}[b]{0.31\linewidth}
\centering
\small
\begin{tabular}{|l|c|c|}
\hline $\tau=25\%n$ & r $\le$ 25\% & r $>$ 25\%\\
\hline
correct & 1134 & 2937  \\
incorrect & 276 & 136 \\
\hline
\end{tabular}
(a)
\end{minipage}
\hspace{0.01cm}
\begin{minipage}[b]{0.31\linewidth}
\centering
\small
\begin{tabular}{|l|c|c|}
\hline $\tau=50\%n$ & r $\le$ 50\% & r $>$ 50\%\\
\hline
correct & 2071 & 2000  \\
incorrect & 331 & 81 \\
\hline
\end{tabular}
(b)
\end{minipage}
\hspace{0.01cm}
\begin{minipage}[b]{0.305\linewidth}
\centering
\small
\begin{tabular}{|l|c|c|}
\hline $\tau$=flexible & label=0 & label=1\\
\hline
correct & 805 & 3266  \\
incorrect & 333 & 79 \\
\hline
\end{tabular}
(c)
\end{minipage}
\vspace{-0.15in}
\caption{Confusion Matrices for (a) using heuristic $h_1$ with $\tau=25\%n$, (b) heuristic $h_1$ with $\tau=50\%n$ and (c) heuristic $h_2$ with flexible $\tau$.}
\label{tab:confusion}
\vspace{-0.23in}
\end{table}
\begin{table}[h!]
\begin{center}
\small
\begin{tabular}{l|c|c|c|c}
\hline \bf Models & Heuristic & Depth ($\tau$) & Beam Size (b) & In-KB acc \% \\
\hline
Global + LDS & $h_1$ & $\tau=25\%n$ & 1 &  93.88 \\
Global + LDS & $h_1$ & $\tau=25\%n$ & 5 &  94.12 \\
Global + LDS & $h_1$ & $\tau=50\%n$ & 5 & 94.23 \\
Global + LDS & $h_2$ & flexible & 5 & 94.44 \\
\hline
\end{tabular}
\end{center}
\vspace{-0.15in}
\caption{\label{tab:HC} CoNLL 2003 Test-b}
\vspace{-0.15in}
\end{table}

In Table ~\ref{tab:HC}, we report the performance of our model on CoNLL 2003 with different choices for the heuristic, search depth, and beam size. The results show that using a beam of size 5 improves upon single greedy search, and using heuristic $h_2$  with flexible depth gives the best performance in terms of both  prediction accuracy and efficiency (due to smaller search trees). When $h_1$ is used, doubling the depth of the search tree brings about only a marginal improvement in accuracy at the cost of doubling the search depth and thus the prediction time.

\subsubsection{Performance Analysis Based on Entity Rarity} 
\begin{figure}[ht]
\centering
\includegraphics[width=0.75\textwidth,height=0.32\textwidth]{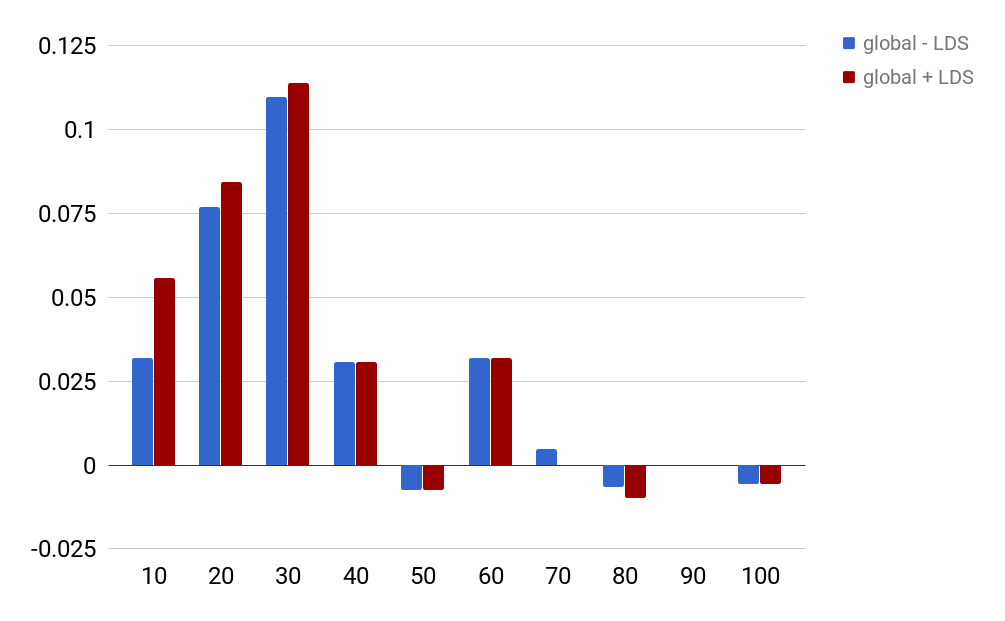}
\vspace{-0.15in}
\caption{\label{fig:gc} Global accuracy minus local accuracy per bin of rarity measure $p(e|m)$ which is scaled to (0, 100) for two coherent global models; model without LDS and model with LDS.}
\vspace{-0.20in}
\end{figure}

We also analyze the behavior of the coherent global models based on the rarity of the entities for the given mention. Specifically, we measure the rarity of $e$ for given query mention $m$ by $p(e|m)$, as defined in \ref{subsec:prior}, scaled to (0, 100). To this end we quantize the value of rarity measure into different bins. For each bin we compute the difference of accuracy between a coherent global model and the local model. Two coherent models are considered here; a global model without LDS (-LDS + conv.global prop in Table~\ref{tab:TAC_F}) and global model with LDS using $h_2$. Figure ~\ref{fig:gc} shows the difference of accuracy between the global and local models per bin. As shown in this figure both global models achieve gains mostly on bins with small $p(e|m)$, which are related to the mentions with rare true entity. Additionally, the impact of the global model when LDS is used is more significant, especially for the mentions whose true entities are most rare according to $p(e|m)$.

\section{Related Work}
Deep learning has been leveraged in recent local and global models. In local models \cite{l:sun} use neural tensor networks to model mention, context and entity embeddings. \cite{rel:hoffman} develop an attention based model to weigh the importance of the words in the query document. The model by \cite{rel:ibm} utilizes neural tensor network, multi-perspective cosine similarity and lexical composition and decomposition. In global models, the early models \citep{l:Milne, l:Ferragina} decomposes the problem over mentions. \cite{l:hoffart} use an iterative heuristic to prune edges among mention and entity. \cite{l:Cheng} use an integer linear program solver and \cite{l:Ratinov} apply SVM to use relation scores as ranking features.

In recent global models, Personalized PageRank (PPR) \citep{l:Jeh} is adopted by several studies~\citep{l:Han,l:He,l:Alhelbawy,l:pershina}. \cite{l:yamada} extend the skip-gram model \citep{l:Mikolov} to learn the relatedness of entities using the linking structure of the KB. \cite{rel:hoffman} use a Conditional Random Field (CRF) based model to capture the interrelationship among entities in the same document whereas \cite{rel:google} introduce a soft k-max attention model to weight the importance of other entities in the document in making prediction for any given query.

In our proposed global model we address the intractable global optimization in a search framework. Specifically we use Limited Discrepancy Search \citep{JMLR:v15:doppa14a}. Initialized with a local solution, LDS only explores a small number corrections to the hard queries. The propagation of the corrections enables the correction of other mistakes (even those with high local confidence) and allows us to reach high quality solutions with shallow searches. Moreover, the heuristic to prioritize the discrepancies can noticeably reduce the time without hurting the performance.
\section{Conclusions} In this paper we study the problem of entity disambiguation. We are inspired by the observation that local models in this task tend to produce reasonable solutions, such that with only a small number of corrections and a proper propagation of  these corrections throughout the document, one can quickly find superior solutions that are globally more coherent.

Based on this observation, we propose a search based approach that starts from an initial solution from a local model, and uses Limited Discrepancy Search (LDS) to search through the space of possible corrections with the goal of improving the linking performance. The experimental results show that our global model improves the state of the art on both the CoNLL 2003 and TAC 2010 benchmarks. For future research, we are interested in further understanding of the strengths and weaknesses of the LDS-based approach for different types of queries and entities. We are also interested in applying different local models to initialize the LDS search.

% include your own bib file like this:
%\bibliographystyle{acl}
%\bibliography{acl2018}
\bibliographystyle{acl_natbib}
\bibliography{acl2018}

\begin{thebibliography}{25}
\expandafter\ifx\csname natexlab\endcsname\relax\def\natexlab#1{#1}\fi

\bibitem[{Alhelbawy and Gaizauskas(2014)}]{l:Alhelbawy}
Ayman Alhelbawy and Robert Gaizauskas. 2014.
\newblock \href {http://www.aclweb.org/anthology/P14-2013} {Graph ranking for
  collective named entity disambiguation}.
\newblock \emph{In Proc. 52nd Annual Meeting of the Association for
  Computational Linguistics, ACL}.

\bibitem[{Cheng and Roth(2013)}]{l:Cheng}
Xiao Cheng and Dan Roth. 2013.
\newblock \href {http://cogcomp.org/papers/ChengRo13.pdf} {Relational inference
  for wikification}.
\newblock \emph{In Proc. of EMNLP}.

\bibitem[{Doppa et~al.(2014)Doppa, Fern, and Tadepalli}]{JMLR:v15:doppa14a}
Janardhan~Rao Doppa, Alan Fern, and Prasad Tadepalli. 2014.
\newblock \href {http://jmlr.org/papers/v15/doppa14a.html} {Structured
  prediction via output space search}.
\newblock \emph{Journal of Machine Learning Research}, 15:1317--1350.

\bibitem[{Ferragina and Scaiella(2010)}]{l:Ferragina}
Paolo Ferragina and Ugo Scaiella. 2010.
\newblock \href
  {https://pdfs.semanticscholar.org/b1a1/3b7c0911eb06fc347f77c18e1f6fdf8f1fb4.pdf}
  {Tagme: on-the-fly annotation of short text fragments (by wikipedia
  entities)}.
\newblock \emph{In Proc. of the 19th ACM International Conference on
  Information Knowledge and Management, CIKM}.

\bibitem[{Francis-Landau et~al.(2016)Francis-Landau, Durrett, and
  Klein}]{l:Francis}
Matthew Francis-Landau, Greg Durrett, and Dan Klein. 2016.
\newblock \href {https://arxiv.org/abs/1604.00734} {semantic similarity for
  entity linking with convolutional neural networks}.
\newblock \emph{Annual Conference of the North American Chapter of the
  Association for Computational Linguistics, NAACL}.

\bibitem[{Ganea and Hofmann(2017)}]{rel:hoffman}
Octavian-Eugen Ganea and Thomas Hofmann. 2017.
\newblock \href {https://arxiv.org/pdf/1704.04920.pdf} {Deep joint entity
  disambiguation with local neural attention}.
\newblock \emph{In Proc. of Empirical Methods in Natural Language Processing}.

\bibitem[{Globerson et~al.(2016)Globerson, Lazic, Chakrabarti, Subramanya,
  Ringgaard, and Pereira}]{rel:google}
Amir Globerson, Nevena Lazic, Soumen Chakrabarti, Amarnag Subramanya, Michael
  Ringgaard, and Fernando Pereira. 2016.
\newblock \href {http://www.aclweb.org/anthology/P16-1059} {Collective entity
  resolution with multi-focal attention}.
\newblock \emph{In Proc. of Association for Computational Linguistics, ACL}.

\bibitem[{Han and Sun(2011)}]{l:Han}
Xianpei Han and Le~Sun. 2011.
\newblock \href {https://www.aclweb.org/anthology/P/P11/P11-1095.pdf} {A
  generative entity-mention model for linking entities with knowledge base}.
\newblock \emph{In Proc. of the 49th Annual Meeting of the Association for
  Computational Linguistics: Human Language Technologies, ACLHLT}.

\bibitem[{He et~al.(2013)He, Liu, Li, Zhou, Zhang, and Wang}]{l:He}
Zhengyan He, Shujie Liu, Mu~Li, Ming Zhou, Longkai Zhang, and Houfeng Wang.
  2013.
\newblock \href {http://www.aclweb.org/anthology/P13-2006} {Learning entity
  representation for entity disambiguation}.
\newblock \emph{Annual Meeting of the Association for Computational
  Linguistics: System Demonstrations, ACL}.

\bibitem[{Heng et~al.(2015)Heng, Joel, and Ben}]{tac:ji15}
Ji~Heng, Nothman Joel, and Hachey Ben. 2015.
\newblock \href {http://nlp.cs.rpi.edu/paper/kbp2015.pdf} {Overview of
  tac-kbp2015 tri-lingual entity discovery and linking}.
\newblock \emph{In Proc. Text Analysis Conference, TAC}.

\bibitem[{Hoffart et~al.(2013)Hoffart, Berberich, and Weikum}]{l:hoffart13}
Johannes Hoffart, Klaus Berberich, and Gerhard Weikum. 2013.
\newblock \href
  {https://people.mpi-inf.mpg.de/~kberberi/publications/2010-mpii-tra.pdf} {A
  spatially and temporally enhanced knowledge base from wikipedia; yago2}.
\newblock \emph{Artificial Intelligence}.

\bibitem[{Hoffart et~al.(2011)Hoffart, Yosef, Bordino, and etc.}]{l:hoffart}
Johannes Hoffart, Mohamed~Amir Yosef, Ilaria Bordino, and etc. 2011.
\newblock \href {http://www.aclweb.org/anthology/D11-1072} {Robust
  disambiguation of named entities in text}.
\newblock \emph{In Proc. of Empirical Methods in Natural Language Processing,
  EMNLP}.

\bibitem[{Jeh and Widom(2003)}]{l:Jeh}
Glen Jeh and Jennifer Widom. 2003.
\newblock \href {http://infolab.stanford.edu/~glenj/spws.pdf} {Scaling
  personalized web search}.
\newblock \emph{In Proceedings of the 12th international conference on World
  Wide Web, pages 271–279. ACM}.

\bibitem[{Ji et~al.(2010)Ji, Grishman, Dang, Griffitt, and Ellis}]{l:ji10}
Heng Ji, Ralph Grishman, Hoa~Trang Dang, Kira Griffitt, and Joe Ellis. 2010.
\newblock \href
  {https://pdfs.semanticscholar.org/b7fb/11ef06b0dcdc89ef0a5507c6c9ccea4206d8.pdf}
  {Overview of the tac 2010 knowledge base population track}.
\newblock \emph{In Proc. of the 3rd Text Analysis Conference, TAC}.

\bibitem[{Ji et~al.(2014)Ji, Joel, and Ben}]{tac:ji14}
Heng Ji, Nothman Joel, and Hachey Ben. 2014.
\newblock \href {https://www3.nd.edu/~qzhi/papers/edl2014overview.pdf}
  {Overview of tac-kbp2014 entity discovery and linking tasks}.
\newblock \emph{In Proc. Text Analysis Conference, TAC}.

\bibitem[{Lev-Arie~Ratinov and Anderson(2011)}]{l:Ratinov}
Doug~Downey Lev-Arie~Ratinov, Dan~Roth and Mike Anderson. 2011.
\newblock \href {https://web.eecs.umich.edu/~mrander/pubs/RatinovDoRo.pdf}
  {Local and global algorithms for disambiguation to wikipedia}.
\newblock \emph{In Proc. of the 49th Annual Meeting of the Association for
  Computational Linguistics: Human Language Technologies, ACL HLT}.

\bibitem[{Manning and McClosky(2014)}]{l:corenlp}
Surdeanu Mihai Bauer John Finkel Jenny Bethard Steven~J. Manning,
  Christopher~D. and David McClosky. 2014.
\newblock \href {https://stanfordnlp.github.io/CoreNLP/citing.html} {The
  stanford corenlp natural language processing toolkit}.
\newblock \emph{Annual Meeting of the Association for Computational
  Linguistics: System Demonstrations, ACL}, pages 55--60.

\bibitem[{Mikolov et~al.(2013)Mikolov, Sutskever, Chen, Corrado, and
  Dean}]{l:Mikolov}
Tomas Mikolov, Ilya Sutskever, Kai Chen, Greg~S. Corrado, and Jeff Dean. 2013.
\newblock \href
  {https://papers.nips.cc/paper/5021-distributed-representations-of-words-and-phrases-and-their-compositionality.pdf}
  {Distributed representations of words and phrases and their
  compositionality}.
\newblock \emph{In Advances in Neural Information Processing Systems, NIPS}.

\bibitem[{Milne and Witten(2008)}]{l:Milne}
David~N. Milne and Ian~H. Witten. 2008.
\newblock \href
  {https://www.semanticscholar.org/paper/Learning-to-link-with-wikipedia-Milne-Witten/07abd02f02774d178f26ca99937e5f94001a9ec9}
  {Learning to link with wikipedia}.
\newblock \emph{In Proc. of the 17th ACM Conference on Information and
  Knowledge Management, CIKM}.

\bibitem[{Nevena~Lazic and Pereira(2015)}]{l:lazic}
Michael~Ringgaard Nevena~Lazic, Amarnag~Subramanya and Fernando Pereira. 2015.
\newblock \href {https://www.aclweb.org/anthology/Q/Q15/Q15-1036.pdf} {Plato: A
  selective context model for entity resolution}.
\newblock \emph{Transactions of the Association for Computational Linguistics}.

\bibitem[{Pershina et~al.(2015)Pershina, He, and Grishman}]{l:pershina}
Maria Pershina, Yifan He, and Ralph Grishman. 2015.
\newblock \href {http://www.aclweb.org/anthology/N15-1026} {Personalized page
  rank for named entity disambiguation}.
\newblock \emph{Annual Conference of the North American Chapter of the
  Association for Computational Linguistics, NAACL}.

\bibitem[{Sil and Florian(2016)}]{l:sil:16}
Avirup Sil and Radu Florian. 2016.
\newblock \href {https://arxiv.org/pdf/1712.01797.pdf} {Towards language
  independent named entity linking}.
\newblock \emph{Annual Meeting of the Association for Computational
  Linguistics: System Demonstrations, ACL}.

\bibitem[{Sil et~al.(2018)Sil, Kundu, Florian, and Hamza}]{rel:ibm}
Avirup Sil, Gourab Kundu, Radu Florian, and Wael Hamza. 2018.
\newblock \href {https://arxiv.org/pdf/1712.01813.pdf} {Neural cross-lingual
  entity linking}.
\newblock \emph{Thirty-Second AAAI Conference on Artificial Intelligence,
  AAAI}.

\bibitem[{Sun et~al.(2015)Sun, Lin, Tang, Yang, Ji, and Wang}]{l:sun}
Yaming Sun, Lei Lin, Duyu Tang, Nan Yang, Zhenzhou Ji, and Xiaolong Wang. 2015.
\newblock \href {https://www.ijcai.org/Proceedings/15/Papers/192.pdf} {Modeling
  mention, context and entity with neural networks for entity disambiguation}.
\newblock \emph{International Joint Conference on Artificial Intelligence,
  IJCAI}.

\bibitem[{Yamada et~al.(2016)Yamada, Shindo, Takeda, and Takefuji}]{l:yamada}
Ikuya Yamada, Hiroyuki Shindo, Hideaki Takeda, and Yoshiyasu Takefuji. 2016.
\newblock \href {http://www.aclweb.org/anthology/K16-1025} {Joint learning of
  the embedding of words and entities for named entity disambiguation}.
\newblock \emph{International Conference on Computational Linguistics, COLING}.

\end{thebibliography}
\appendix
\end{document}